%% file: paper_ICDM.tex
\documentclass[conference]{IEEEtran}

\usepackage[noadjust]{cite}
\usepackage{amsmath,amssymb,amsfonts}
\usepackage{graphicx}
\usepackage{textcomp}
\usepackage{xcolor}

\usepackage{hyperref}
\usepackage{caption}
\usepackage{subcaption}
\usepackage{algorithm}
\usepackage[noend]{algpseudocode}
\usepackage{algorithmicx}
\usepackage{multirow}
\usepackage{multicol}
\usepackage{booktabs}       % professional-quality tables

\usepackage{bm}
\usepackage{pifont} % symbols font

\usepackage{scrextend}
\usepackage{fancyhdr}

\IEEEoverridecommandlockouts
\newcommand{\eg}{\textit{e.g.}}
\newcommand{\ie}{\textit{i.e.}}

\DeclareMathOperator*{\argmax}{arg\,max}

\def\algbackskip{\hskip-\ALG@thistlm}

\newcommand\MyBox[2]{
  \fbox{\lower0.75cm
    \vbox to 1.7cm{\vfil
      \hbox to 1.7cm{\hfil\parbox{1.4cm}{#1\\#2}\hfil}
      \vfil}%
  }%
}

\newcommand\MyBoxWide[2]{
  \fbox{\lower0.75cm
    \vbox to 1.7cm{\vfil
      \hbox to 3cm{\hfil\parbox{2.6cm}{#1\\#2}\hfil}
      \vfil}%
  }%
}

\newcommand{\KwInput}{\hspace*{\algorithmicindent} \textbf{Input:}}

\begin{document}

\title{Promoting Fairness through\\ Hyperparameter Optimization
\thanks{The project CAMELOT (reference POCI-01-0247-FEDER-045915) leading to this work is co-financed by COMPETE 2020, NORTE 2020 and by the Portuguese Foundation for Science and Technology - FCT under the CMU-Portugal international partnership.}
}

\author{

\IEEEauthorblockN{
\hfill% \hspace*{\fill}
Andr\'e F. Cruz\IEEEauthorrefmark{1}
\hspace{1em}
Pedro Saleiro\IEEEauthorrefmark{1}
\hspace{1em}
Catarina Bel\'em\IEEEauthorrefmark{1}
\hspace{1em}
Carlos Soares\IEEEauthorrefmark{2}
\hspace{1em}
Pedro Bizarro\IEEEauthorrefmark{1}
\hfill%\hspace*{\fill}
}

\IEEEauthorblockA{\IEEEauthorrefmark{1}\textit{Feedzai} \hspace{2em}
% \IEEEauthorrefmark{2}\textit{Fraunhofer AICOS} \hspace{1em}
\IEEEauthorrefmark{2}\textit{Universidade do Porto}\\
\{andre.cruz, pedro.saleiro, catarina.belem, pedro.bizarro\}@feedzai.com \hspace{1em} csoares@fe.up.pt
}
}

\maketitle

\begin{abstract}
Considerable research effort has been guided towards algorithmic fairness but real-world adoption of bias reduction techniques is still scarce.
Existing methods are either metric- or model-specific, require access to sensitive attributes at inference time, or carry high development or deployment costs.
This work explores the unfairness that emerges when optimizing ML models solely for predictive performance, and how to mitigate it with a simple and easily deployed intervention: fairness-aware hyperparameter optimization (HO).
%
%This work explores the unfairness that emerges from traditional ML model development, and how to mitigate it with a simple and easily deployed intervention: fairness-aware hyperparameter optimization (HO).
%
We propose and evaluate fairness-aware variants of three popular HO algorithms: Fair Random Search, Fair TPE, and Fairband.
%
%Our method enables practitioners to adapt pre-existing business operations to accommodate fairness objectives in a frictionless way and with controllable fairness-performance trade-offs.
% Additionally, it can be coupled with existing bias reduction techniques to tune their hyperparameters.
%
We validate our approach on a real-world bank account opening fraud case-study, as well as on three datasets from the fairness literature.
Results show that, without extra training cost, it is feasible to find models with 111\% mean fairness increase and just 6\% decrease in performance when compared with fairness-blind HO.
\end{abstract}


\section{Introduction}

\thispagestyle{fancy}
\renewcommand{\headrulewidth}{0pt}
\renewcommand{\footrulewidth}{0pt}
\cfoot{Accepted at the $21^{st}$ IEEE International Conference
on Data Mining (ICDM 2021).}

%Machine Learning (ML) has been increasingly used to aid decision-making in sensitive domains, including healthcare~\cite{Rajkomar2019}, criminal justice~\cite{berk2018fairness}, and financial services~\cite{branco2020interleaved}.
%These algorithmic decision-making systems are accumulating societal responsibilities, often without human oversight. At the same time, Machine Learning (ML) models are usually optimized for a single global metric of predictive performance (\eg, binary cross-entropy loss on the training set), without taking into account possible side-effects and their real-world implications.

Algorithmic bias arises when a Machine Learning (ML) model displays disparate error rates across sub-groups of the population, hurting individuals based on ethnicity, age, gender, or any other sensitive attribute~\cite{Angwin2016,Bartlett2019,Buolamwini2018}. This has several causes, from historical biases encoded in the data, to misrepresented populations in data samples, noisy labels, development decisions (\eg, missing values imputation), or simply the nature of learning under severe class-imbalance~\cite{Suresh2019}.

Despite growing awareness, as of today, most companies are unsure of the cost implications of tackling algorithmic bias, not only from model performance degradation but also from extra development costs (both human and computational).
The current ML landscape lacks practical methodologies and tools to seamlessly integrate fairness objectives and bias reduction techniques in existing real-world ML pipelines~\cite{bigdatasocialsciences}.
As a consequence, treating fairness as a primary objective when developing ML systems is not yet standard practice.

\begin{figure}[t]
    \centering
    \includegraphics[width=\columnwidth]{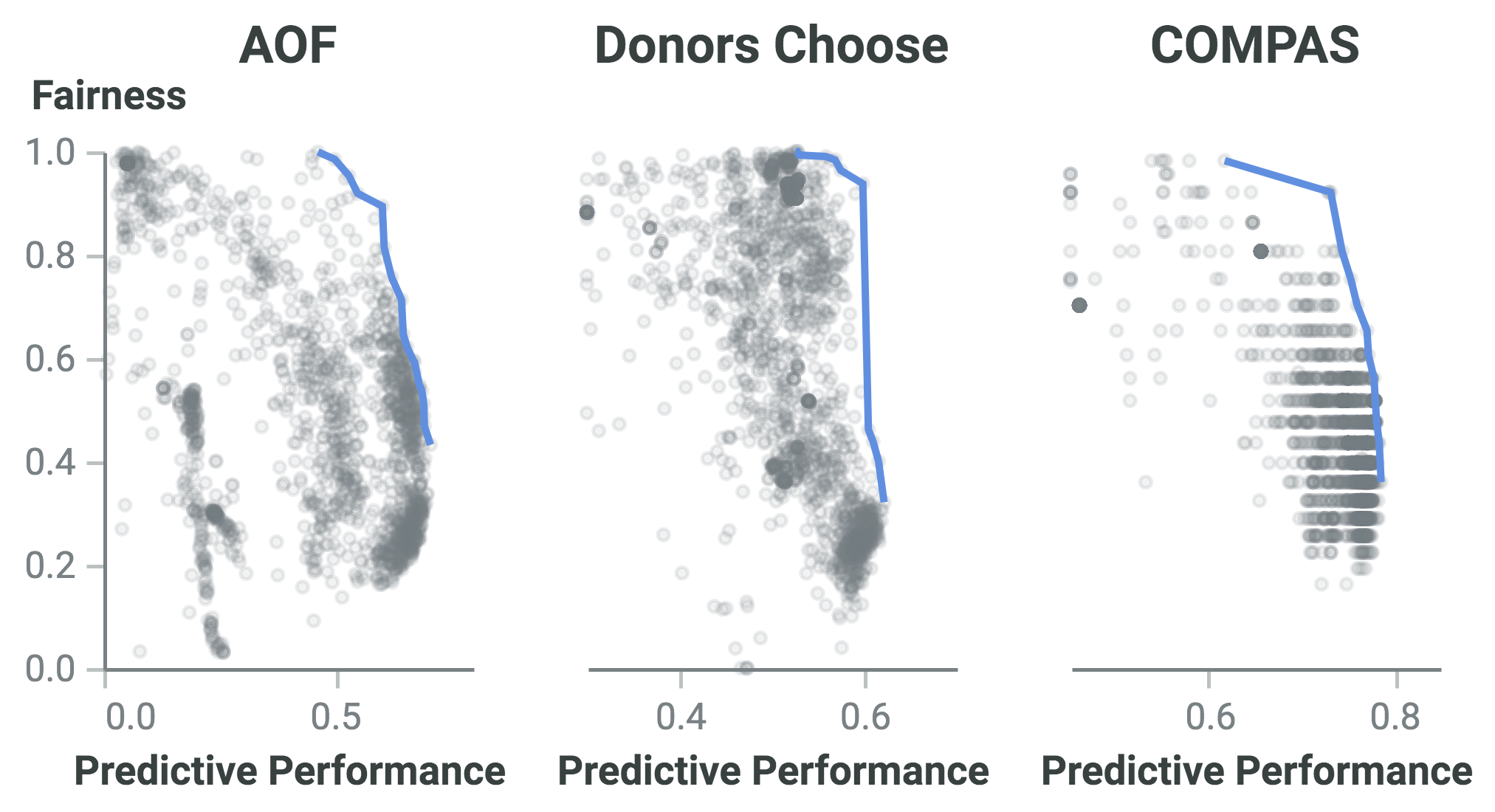}

    \caption{Fairness-performance Pareto frontier (blue line) on the in-house Account Opening Fraud dataset (left), the Donors Choose dataset (middle), and the COMPAS dataset (right).}
    \label{fig:pareto_frontiers}
\end{figure}

% \begin{figure}[tb]
%     \centering
%     \includegraphics[width=1.0\columnwidth]{figures/plots/intro_tradeoff.png}
%     % \includegraphics[width=0.9\columnwidth]{figures_Feedzai_theme/fairness_performance_scatter.png}
%     %\caption{
%     %Fairness-performance trade-off on the Adult dataset.
%     %The top 10\% models with highest predictive performance are identified as the \textit{fairness-blind} region.}
%     \caption{Fairness-performance trade-off on the Adult dataset. Marked with an A, the model with highest performance; marked with a B, a model with 0.8\% lower performance and 44.8\% higher fairness than A, arguably a better trade-off. %In orange, the linear regression relationship between performance and fairness; in the red rectangle, the top 10\% models with highest performance (the target region for a fairness-blind process); in light gray, the fairness-performance Pareto frontier; marked with an A, the model with highest performance; marked with a B, a model with 0.8\% lower performance and 44.8\% higher fairness than A, arguably a better trade-off, and one that would not be found by traditional fairness-blind techniques.
%     }
%     \label{fig:intro_tradeoff}
% \end{figure}

%
Existing bias reduction techniques only target specific stages of the ML pipeline (\eg, data sampling, model training), and often only apply to a single fairness definition or family of ML models~\cite{Zafar2017,Agarwal2018}, limiting their adoption in practice.
Moreover, the absence of major breakthroughs in algorithmic fairness suggests that an exhaustive search over all models and bias reduction techniques is necessary to find optimal trade-offs.
%, hence discouraging ML practitioners.
% Figure~\ref{fig:intro_tradeoff} illustrates this over 10K models trained with different hyperparameter configurations.

% %
% Currently, standard hyperparameter optimization (HO) and model selection processes are fairness-blind, solely optimizing for predictive performance.
% By doing so, these methods unknowingly target models with low fairness (region marked with a red rectangle).
% However, as shown by the plotted Pareto frontier~\cite{Pareto1906}, we can achieve significant fairness improvements at small performance costs.
% For instance, model B achieves 44.8\% higher fairness than model A (the model with highest predictive performance), at a cost of 0.8\% decrease in predictive performance, arguably a better trade-off.
% %
% While current fairness-blind techniques target model A, we aim to target the region of optimal fairness-performance trade-offs to which model B belongs.
% %
% Indeed, we observe a large spread over the fairness metric at any level of predictive performance, even within this fairness-blind region.
% Thus, it is absolutely possible to select fairer hyperparameter configurations without significant decrease in predictive performance. %, the challenge is guiding the search process.
% %

%
This work explores, in the context of a large-scale real-world case-study of account opening fraud, the unfairness that emerges from traditional ML model development, and how to mitigate it with a simple and easily deployed intervention: fairness-aware HO.
%
%With this in mind, we propose to foster algorithmic fairness through fairness-aware HO.
By making the hyperparameter search fairness-aware, %while maintaining resource-efficiency, 
we enable ML practitioners to adapt pre-existing business operations to accommodate fairness with controllable extra cost and little implementation friction.

Figure~\ref{fig:pareto_frontiers} shows the fairness-performance Pareto frontiers~\cite{Pareto1906} obtained from training 10K different hyperparameter configurations of 5 ML algorithms on each of three real-world datasets.
%
% for different datasets (10K models trained on each dataset). % from training 10K models with different hyperparameter configurations, on three datasets.
We observe a large spread over the fairness metric at any level of predictive performance, 
indicating it is clearly possible to select fairer models without a significant performance decrease.
% Thus, it is evidently possible to select fairer hyperparameter configurations without a significant decrease in performance.
%
Based on this insight, we extend three popular hyperparameter tuners (Random Search, TPE~\cite{Bergstra2011} and Hyperband~\cite{Li2016b}) to optimize for both predictive performance and fairness through a weighted scalarization. % controlled by an $\alpha$ parameter.
As budget %(be it time or computational) 
is often limited in real-world scenarios, we give special focus to Fairband (our extension to Hyperband), and propose a heuristic to automatically find an adequate fairness-performance trade-off using this tuner (dubbed FB-auto).

We apply our approach on a real-world bank account opening fraud detection problem.
%
% In account opening fraud, a malicious actor will attempt to open a new bank account using a stolen or synthetic identity (or both), in order to quickly max out its line of credit~\cite{FraudSurvey2016}.
%
When developing ML models to detect fraud, banks optimize for a single metric of predictive performance (\eg, fraud recall). However, as shown in our experiments, the models with highest fraud recall have disparate false positive rates (FPR) on specific groups of applicants.
%
% In this context, a false positive is a legitimate individual that was denied access to a bank account and the accompanying credit line, which are essential services for economic well being and social mobility. %, and even declared a basic right by the European Parliament~\cite{basic_account_EU}.
In this context, a false positive is a legitimate individual that was denied access to a bank account and its line of credit, an essential service for economic well being and social mobility~\cite{basic_account_EU}.
Therefore, we consider a model to be fair if the probability of a legitimate applicant being denied access to a bank account is independent from their membership to a protected group (\eg, FPR parity across age groups).

By applying our fairness-aware HO strategies we were able to find models with 111\% improved fairness and just 6\% drop in fraud recall when compared to standard HO. % methods.

The summary of our contributions are as follows:
\begin{itemize}

    \item An approach for promoting model fairness that can be easily plugged into current ML pipelines with no extra development or computational cost.

    \item A set of competitive fairness-aware HO algorithms for multi-objective optimization (MOO) of the fairness-performance trade-off that are agnostic to both the explored hyperparameter space and the objective metrics. %(described in Section~\ref{chap:method}).

    \item A heuristic to automatically set the fairness-performance trade-off parameter. %(described in Section~\ref{sec:dynamic_alpha}).

    % \item Strong empirical evidence that HO is an effective way to navigate the fairness-performance trade-off.

    \item Competitive results on a real-world fraud detection case-study, as well as on three public datasets. %: fair HO achieves significantly improved fairness at a small performance cost, and no extra budget when compared to literature HO baselines.

\end{itemize}

% %%%%%%%%%%%%%%%%%%%%%%%%%%%%%%%%%%%%%%%%%%%%%%%%%%%%%%%%%%%%%%%%%%%%%%%%%%%%%%%%%%%%%%%%%%%%%%%%%%%%%%%%%%%%%%%%
% % 
% %                                                  RELATED WORK 
% % 
% % %%%%%%%%%%%%%%%%%%%%%%%%%%%%%%%%%%%%%%%%%%%%%%%%%%%%%%%%%%%%%%%%%%%%%%%%%%%%%%%%%%%%%%%%%%%%%%%%%%%%%%%%%%%%%

\section{Related Work}
\label{sec:related_work}

Algorithmic fairness research work can be broadly divided into three families: pre-processing, in-processing, and post-processing.

\textbf{Pre-processing} methods aim to improve fairness before the model is trained, by modifying the input data such that it no longer exhibits biases.
The objective is often formulated as learning new representations that are invariant to changes in specified factors (\eg, membership in a protected group)~\cite{calmon2017optimized,Creager2019,Edwards2016,Zemel2013}.
However, by acting on the data itself, and in the beginning of the ML pipeline, fairness may not be guaranteed on the end model that will be used in the real-world.

\textbf{In-processing} methods alter the model's learning process in order to penalize unfair decision-making.
The objective is often formulated as optimizing predictive performance under fairness constraints (or optimizing fairness under predictive performance constraints)~\cite{Agarwal2018,Zafar2017}.
Another approach is optimizing for complex predictive performance metrics which include some fairness notion~\cite{Bechavod2017}, akin to regularization.
However, these approaches are highly model- and metric-dependent, and even non-existent for numerous state-of-the-art ML algorithms.

\textbf{Post-processing} methods aim to adjust an already trained classifier such that fairness constraints are fulfilled.
This is usually done by calibrating the decision threshold~\cite{Fish2016,Hardt2016}.
%As such, these approaches are flexible and applicable to any score-based classifier.
%
However, one may argue that, by acting on the model \textit{after} it was learned, this process is inherently sub-optimal~\cite{Woodworth2017}. It is akin to knowingly learning a biased model and then correcting these biases, instead of learning an unbiased model from the start.
Moreover, these approaches require production-time access to sensitive features, forcing companies to ask all clients their ethnicity, religion, gender, or any other feature for fairness criteria. %Individuals are rightfully wary of the implications such data collection could have.    %% TODO: citation needed (?)

%% ------------ -----------------
%% HO
%% ------------ -----------------
% Numerous methods have been proposed to address HO, including the traditionally employed algorithms, \eg, random-search, and recently introduced resource-aware algorithms, \eg, Successive Halving~\cite{Karnin2013} and Hyperband~\cite{Li2016b}.

Although a largely unexplored direction, algorithmic fairness can also be tackled from an HO perspective.
%
% Numerous methods have been proposed to address HO, including the traditionally employed algorithms, \eg, random-search, and recently introduced resource-aware algorithms, \eg, Successive Halving~\cite{Karnin2013} and Hyperband~\cite{Li2016b}.
%

\textbf{Random Search} (RS), one of the simplest and most flexible HO methods, iteratively selects combinations of random hyperparameter values and trains them on the full training set until the allocated budget is exhausted.
% Each hyperparameter configuration is trained on the full training set, and the best performing configuration is selected by evaluating on the validation set.
%
Although simple in nature, RS has several advantages that keep it relevant nowadays, including having no assumptions on the hyperparameter space, on the objective function, or even on the allocated budget (\eg, it may run indefinitely).
%Additionally, 
RS is known to generally perform better than grid search~\cite{Bergstra2012} and to converge to the optimum as budget increases.

\textbf{Bayesian Optimization} (BO) is a state-of-the-art HO approach that consists in placing a prior (usually a Gaussian process (GP)) over the objective function to capture beliefs about its behavior~\cite{Shahriari2016}.
% It iteratively updates the \textit{prior} distribution, $p(\lambda)$, using the observed goal, $Y$, and forms a \textit{posterior} distribution of its behavior, $p(\lambda|Y)$.
% % It iteratively updates the \textit{prior} distribution, $p(y)$, using the observed loss in the evidence \textit{trial}, $y$, and forms a \textit{posterior} distribution of its behavior, $p(y|\lambda)$.
% %
% Afterwards, an acquisition function $a$ is constructed to determine the next query point $\lambda^{(i+1)}=\argmax_{\lambda}{a(\lambda)}$. This process is repeated in a sequential manner, continuously improving the approximation of the underlying objective function~\cite{Hutter2011a}.
%
%% Constrained BO %%
Previous work has extended BO to constrained optimization~\cite{cBO2014Kusner,cBO2014Snoek}, in which the goal is to optimize a given metric subject to any number of data-dependent constraints.
Recently, Perrone et al.~\cite{perrone2021fair} applied this approach to the fairness setting by weighing the acquisition function by the likelihood of fulfilling the fairness constraints.
However, constrained optimization approaches inherently target a single fairness-performance trade-off (which itself may not be feasible), leaving practitioners unaware of the possible fairness choices and their performance costs.
Additionally, vanilla BO (based on GPs) cannot handle categorical hyperparameters, scales cubically in the number of data points, and performs poorly on high dimensional or conditional search spaces~\cite{Hutter2019}. % Eggensperger2013

%%%%%%%%%
%% TPE
%%%%%%%%%
%Due to its difficulty in handling categorical variables and high-dimensional spaces,
% As such, several alternatives to vanilla BO have been proposed~\cite{Hutter2011a,Bergstra2011,Snoek2015,Springenberg2016}.
%, substituting the GP prior with random embeddings~\cite{Wang2012b}, a random forest~\cite{Hutter2011a}, a neural network~\cite{Snoek2015}, or a Parzen-tree~\cite{Bergstra2011}, among others~\cite{pmlr-v84-wang18c,Springenberg2016}.
%
The Tree-structured Parzen Estimator (TPE)~\cite{Bergstra2011} is a state-of-the-art BO algorithm that uses Parzen windows~\cite{parzen1962} to model the density of hyperparameter choices below and above some goal value, $y^*$, chosen to be some quantile of the observed goals.
%
% The densities are estimated using Parzen windows~\cite{parzen1962}. %and the ratio of densities, $\frac{p(\lambda | y > y^*)}{p(\lambda | y \leq y^*)}$, is used to suggest promising hyperparameter choices.
As the Parzen estimators follow a tree-structure of conditional spaces, TPE shows good performance on high-dimensions and hierarchically-defined hyperparameters. %(\eg, the choice of model hyperparameters depends on the model choice). %~\cite{Bergstra2013}.
Nonetheless, BO has been shown to lag behind bandit-based methods when under tight budget constraints~\cite{Li2016b,Hutter2011a}. %~\cite{Li2016b,Falkner2018b,Hutter2011a}.

\textbf{Successive Halving} (SH)~\cite{Karnin2013,Jamieson2016}
casts the task of HO as identifying the best arm in a multi-armed bandit setting.
Given a budget for each iteration, $B_i$, SH (1) uniformly allocates it to a set of arms (hyperparameter configurations), (2) evaluates their performance, (3) discards the worst half, and repeats from step 1 until a single arm remains.
Thereby, the budget for each surviving configuration is effectively doubled at each iteration.
SH’s key insight stems from extrapolating the rank of configurations’ performances from their rankings on diminished budgets (low-fidelity approximations).
However, SH itself carries two parameters for which there is no clear choice of values: the total budget $B$ and the number of sampled configurations $n$.
%
% As the method leverages performance evaluations on a smaller budget (averaging $B/n$) to infer rank on a higher budget, 
We must consider the trade-off between evaluating a higher number of configurations (higher $n$) on an averaged lower budget per configuration ($B / n$), or evaluating a lower number of configurations (lower $n$) on an averaged higher budget.
The higher the average budget, the more accurate the extrapolated rankings will be, but a lower number of configurations will be explored (and vice-versa).
% Based on a stochastic learning process, SH may not be able to distinguish performance of different runs in the allocated budget for each configuration, which averages $B/n$. For a fixed $B$, the trade-off between choosing a low number of configurations (small $n$) with large average training time, or a higher number of configurations (larger $n$) with smaller training time must be considered. %
SH's performance has been shown to compare favorably to several competing bandit strategies from the literature~\cite{Audibert2010}.   %\cite{Audibert2010,Even-Bar2006,Jamieson2014,Kalyanakrishnan2012}.

\textbf{Hyperband} (HB)~\cite{Li2016b} addresses this ``$n$ \textit{versus} $B/n$'' trade-off by splitting the total budget into different instances of the trade-off, then calling SH as a subroutine for each one.
This is essentially a grid search over feasible values of $n$.
HB takes two parameters $R$, the maximum amount of resources allocated to any single configuration; and $\eta$, the ratio of budget increase in each SH round ($\eta=2$ for the original SH).
Each SH run, dubbed a bracket, 
%(each corresponding to a different value of $s$), 
is parameterized by the number of sampled configurations $n$, and the minimum resource units allocated to any configuration $r$.
The algorithm features an outer loop that iterates over possible combinations of ($n$,$r$), and an inner loop that executes SH with the aforementioned parameters fixed.
The outer loop is executed $s_{max}+1$ times, $s_{max}=\lfloor log_\eta(R) \rfloor$, and the inner loop (SH) takes approximately $B$ resources.
Thus, the execution of Hyperband takes a budget of $(s_{max}+1)B$.
Table~\ref{tab:hyperband_brackets} displays the number of configurations and budget per configuration within each bracket when considering $\eta=3$ and $R=100$.

\begin{table}[t]    % Keep this table at the top of a page, it's important for future reference!
    \caption{Number of sampled configurations, $n_i$, and budget per configuration, $r_i$, per HB iteration, $i$, on each bracket, $s$ (for $\eta=3, R=100$).
    % (when the ratio of budget increase $\eta = 3$, and the maximum budget per configuration $R=100$). %One budget unit equals 1\% of the train dataset.
    }
    \begin{center}
    % \resizebox{\columnwidth}{!}{
    \input{tables/hyperband_brackets}
    % }
    \label{tab:hyperband_brackets}
    \end{center}
\end{table}

%
% In practice, the choice of which bias reduction method to employ is usually done by so-called ``graduate student descent''.
% In practice, a large selection of bias reduction methods has to be evaluated in order to find optimal fairness-performance trade-offs.
%
HO is simultaneously model independent, metric independent, and already an intrinsic component in existing real-world ML pipelines. On the other hand, 
current bias reduction methods either 
(1) act on the input data and may not be able to guarantee fairness on the end model, 
(2) act on the model's training phase and can only be applied to specific model types and fairness metrics, %thus are inflexible with regards to the chosen model or metric,
or (3) act on the model's predictions thus requiring test-time access to sensitive attributes and being limited to act on a sub-optimal space~\cite{Woodworth2017}.
%
% HO can supplement current bias reduction methods by including these in the search space and automatically selecting which method to employ.
%
Therefore, by introducing fairness objectives on the HO phase, we aim to help real-world practitioners to find optimal fairness-performance trade-offs in an easily pluggable manner, irrespective of the underlying model type or bias reduction method.

%%%%%%%%%%%%%%%%%%%%%%%%%%%%%%%%%%%%%%%%%%%%%%%%%%%%%%%%%%%%%%%%%%%%%%%%%%%%%%%%%%%%%%%%%%%%%%%%%
% 
%                                       METHOD / PROPOSED SOLUTION 
% 
%%%%%%%%%%%%%%%%%%%%%%%%%%%%%%%%%%%%%%%%%%%%%%%%%%%%%%%%%%%%%%%%%%%%%%%%%%%%%%%%%%%%%%%%%%%%%%%%%
\section{Approach}
\label{chap:method}

%Our goal is to enable promoting fairness in a flexible and easily pluggable manner, as reaching a large audience of ML practitioners and for-profit organizations requires minimizing attrition.
Persuading a large audience of ML practitioners and for-profit organizations to adopt fairness considerations into their ML pipeline requires minimizing implementation attrition.
We address this issue through fairness-aware HO.
% Particularly, we propose a flexible and off-the-shelf approach that can promptly be integrated into real-world ML standard practices.
% We propose tackling this problem through fairness-aware HO.
%
% By acting on the algorithms' hyperparameter space, we benefit from the advantages of in-processing bias reduction methods, while avoiding its shortcomings (\eg, dependency on the metrics and model).
%
% That is, we aim to improve model fairness by acting on the models' training phase, instead of simply correcting the data (pre-processing), or correcting the predictions (post-processing).
By acting on the algorithms' hyperparameter space, we promote fairness without fundamentally changing the pipeline (\ie, the set of operators used to generate the model) and without changing the methods themselves (\ie, standard off-the-shelf learning algorithms and pre-processing methods are used), providing a flexible and off-the-shelf approach that can promptly be integrated into real-world ML standard practices.

As a case in point, our use-case is of a fintech company that leverages ML solutions to prevent financial fraud. The deployment of an ML model to a client is preceded by a heavy model selection stage where hundreds of models are trained and evaluated under some predetermined performance metric, defined according to business requirements (\eg, maximize recall at 3\% FPR). It is often the case that clients also demand for specific state-of-the-art models including deep learning or boosting-based models~\cite{LGBM}. Accommodating fairer practices in such a complex setting poses many implementation challenges, including the integration of model-specific bias mitigation methods (which may not always comply with business demands) 
%, the unavailability of specific data at production time (\eg sensitive information about the client), among others.
and guaranteeing low-latency requirements are met after model deployment (bias mitigation techniques often incur extra train and evaluation time).
% Sell our solution 
Fairness-aware HO emerges as a seamless and flexible approach that allows decision-makers to have a finer control over which models fulfil the business requirements (both in terms of fairness, predictive performance, and latency requirements).
% In the case of Feedzai, a large percentage of fraud detection models in production are LightGBMs~\cite{LGBM}, for which, to the best of our knowledge, there is no in-processing bias reduction method.
% These models carry several advantages, one of which being their fast train and prediction times, a requirement to comply with our low-latency decision standards.
% As such, the use of fairness-aware HO further fulfills this business requirement.

%%%%%%%%%%%%%
%% MOO Pareto Introduction
%%%%%%%%%%%%%
\subsection{Optimizing for Fairness}
% \subsection{Penalizing Unfairness}

The joint maximization of predictive performance and fairness is an MOO problem, defined as:

\begin{equation}
    \argmax_{\lambda \in \Lambda} G(\lambda) = (\rho(\lambda), \phi(\lambda)) \, ,
\end{equation}
where $\lambda$ is a hyperparameter configuration drawn from the hyperparameter space $\Lambda$, $\rho\colon \Lambda \mapsto \left[0, 1\right]$ is the predictive performance metric, and $\phi\colon \Lambda \mapsto \left[0, 1\right]$ is the fairness metric.
In this context, there is no single optimal solution, but a set of Pareto optimal solutions. A solution $\lambda^*$ is Pareto optimal if no other solution improves on an objective without sacrificing another objective. %, \ie, for the utility function $F$, there is no solution $\lambda$ that fulfills
% \begin{equation}
%     \forall g \in F, \exists h \in F, g(\lambda) \geq g(\lambda^*) \wedge h(\lambda) > h(\lambda^*) .
% \end{equation}
The set of all Pareto optimal solutions is dubbed the Pareto frontier~\cite{Pareto1906}.

MOO approaches generally rely on either Pareto-dominance methods or decomposition methods~\cite{miettinen2012nonlinear}.
The former uses Pareto-dominance relations to impose a partial ordering in the population of solutions.
However, the number of incomparable solutions can quickly dominate the size of the population (the number of sampled hyperparameter configurations). This is further exacerbated for high-dimensional problems~\cite{giagkiozis2015methods,Schutze2011}.
On the other hand, decomposition-based methods employ a \textit{scalarizing} function to reduce all objectives to a single scalar output, inducing a total ordering over all possible solutions.
A popular choice is the weighed $l_p$-norm:
%
%% Although Pareto-based methods do lead to a more comprehensive exploration of the whole Pareto frontier, ...

\begin{equation}
\label{eq:l_p_norm}
    \argmax_{\lambda \in \Lambda} \left\lVert H(\lambda) \right\rVert_p = \left( \sum_{i=1}^k w_i h_i(\lambda)^p \right)^{\frac{1}{p}} ,
\end{equation}
where the weights vector $w$ induces an \textit{a priori} preference over the objectives, 
$H(\lambda) = (h_1(\lambda), \ldots, h_k(\lambda))$, 
and $h_i: \Lambda \mapsto \left[0, 1\right]$. % \forall i \in 1, \ldots, k$.
% and $H\colon \Lambda \mapsto \mathbb{R}^k.$

MOO is notoriously arduous to apply at scale.
However, we can draw key insights from observing that the Pareto frontier geometry in this context is most often \textit{convex} %
% \footnote{
%     In fact, we can obtain a convex Pareto frontier $F_c$ from a non-convex Pareto frontier $F_{\overline{c}}$ by creating a randomized classifier $Q$ from a probability distribution over the set of classifiers in $F_{\overline{c}}$, as $Q$ will be able to access all points in the convex hull (or simplex) of $F_{\overline{c}}$.
% }
(visible in Figure~\ref{fig:pareto_frontiers}).
This simplification of the original problem enables us to employ a decomposition-based method, known to converge faster, and to effectively target solutions on the Pareto frontier.
All Pareto optimal solutions of a convex problem can be obtained by varying the weights vector $w$.
Moreover, 
Giagkiozis and Fleming~\cite{giagkiozis2015methods} demonstrate that the use of $l_p$-norms with a high $p$ value leads to slower convergence, with a steep decrease in the likelihood of finding improved solutions as the search progresses.
As such, %aiming for an aggressive convergence rate at the possible cost of some decreased exploration of the Pareto-frontier, we use the $l_1$-norm. 
we employ the $l_1$-norm, carrying the same guarantees as using $p>1$ for convex problems~\cite{miettinen2012nonlinear}.
%
% This is a popular method for MOO known as \textit{weighted-sum scalarization}~\cite{giagkiozis2015methods,miettinen2012nonlinear}.
% 
Our optimization metric $g$ is defined as:
\begin{equation}
    g(\lambda) = \left\lVert G(\lambda) \right\rVert_1 .
\end{equation}
As we optimize only two goals, we can further define, without loss of generality:
\begin{align}
    \alpha &= w_1 = 1 - w_2 \, , \\
    g(\lambda) &= \alpha \cdot \rho(\lambda) + (1 - \alpha) \cdot \phi(\lambda) \, ,
\label{eq:goal}
\end{align}
where $w_1 = \alpha$ is the relative importance of predictive performance, and $w_2 = 1 - \alpha$ is the relative importance of fairness.
Our task is hence reduced to finding the hyperparameter configuration $\lambda$ from a pre-defined hyperparameter search space $\Lambda$ that maximizes the scalar objective function $g(\lambda)$:
\begin{equation}
    \argmax_{\lambda \in \Lambda} g(\lambda) .
\end{equation}

From an implementation standpoint, all models are evaluated in both fairness and predictive performance metrics on a holdout validation set.
Computing fairness does not incur significant extra computational cost, as it is based on the same predictions used to estimate predictive performance.
Additionally, fairness off-the-shelf assessment libraries are readily available~\cite{Saleiro2018}.

In order to find all Pareto optimal solutions we must vary the weighting parameter $\alpha \in \left[ 0, 1 \right]$. 
Nonetheless, $\alpha$ may indicate some predefined objective preference.
For instance, in a punitive ML setting, an organization may decide it is willing to spend $x$\$ per each $1$ less false positive in the underprivileged class\footnote{In a punitive setting, the advised fairness metric is the ratio of group-wise FPRs, also known as \textit{predictive equality}~\cite{Saleiro2018,Corbett-Davies2017}.}, materializing an explicit fairness-performance trade-off.
If no specific trade-off arises from domain knowledge beforehand, then the set of all Pareto optimal models should be displayed, and the decision on which trade-off to employ should be left to the model's stakeholders.

\subsection{Hyperparameter Tuners}
% \subsection{Fairness-Aware HO}

As seen in Section~\ref{sec:related_work}, the best-suited choice of hyperparameter tuner depends on the task at hand.
RS is the most flexible, carries the least assumptions on the optimization metric, and converges to the optimum as budget increases.
TPE improves convergence speed by attempting to sample only useful regions of hyperparameter space.
Bandit-based methods (\eg, Successive Halving,  Hyperband) are resource-aware, and thus have strong anytime performance, often being the most efficient when under budget constraints.

In this work, we propose to extend three popular hyperparameter tuners to optimize for fairness through a weighted scalarization controlled by an $\alpha$ parameter (default $\alpha=0.5$). We propose fairness-aware variants for RS, TPE, and Hyperband, respectively dubbed \textit{FairRS}, \textit{FairTPE}, and \textit{Fairband}. All of these variants can be easily incorporated in existing ML pipelines at a negligible cost.
As budget (be it time or computational) is seldom unrestricted in real-world projects, we give special focus to Fairband. % will be the focus henceforth.

%\section{Fairband}
Fairband inherently benefits from resource-aware methods' advantages: efficient resource usage, trivial parallelization, as well as being both model- and metric-agnostic.
Furthermore, bandit-based methods are highly exploratory and therefore prone to inspect broader regions of the hyperparameter space.
For instance, in our experiments, HB evaluates approximately six times more configurations than RS with the same budget\footnote{With the parameters used on the HB seminal paper~\cite{Li2016b}, HB evaluates 128 configurations \textit{versus} 21 configurations evaluated by RS on equal budget.}.

By employing a weighted scalarization technique in a bandit-based setting, we rely on the hypothesis that if model $m_a$ represents a better fairness-performance trade-off than model $m_b$ with a short training budget, then this distinction is likely to be maintained with a higher training budget.
Thus, by selecting models based on both fairness and predictive performance, we are guiding the search towards fairer and better performing models.
These low-fidelity estimates of future metrics on lower budget sizes is what drives HB's and SH's efficiency in hyperparameter search.

\subsection{Fairband and Dynamic $\alpha$}
%\subsection{Dynamic $\alpha$}
\label{sec:dynamic_alpha}

Aiming for a complete out-of-the-box experience without the need for specific domain knowledge, we further propose a heuristic for dynamically setting the value of $\alpha$.
With this heuristic our objectives are two-fold: first, we eliminate a hyperparameter that would need specific domain knowledge to be set; second, we promote a wider exploration of the Pareto frontier and a larger variability within the sampled hyperparameter configurations.
We dub the variant of Fairband that employs this heuristic as \textit{FB-auto}.
We note this is not suited for model-based search methods (\eg, BO, TPE), as these rely on a stable optimization metric.

Assuming that $\alpha$ values can indeed guide the search towards different regions of the fairness-performance trade-off (which we will empirically see to be true), our aim is to efficiently explore the Pareto frontier in order to find a comprehensive selection of balanced trade-offs.
As such, if our currently explored trade-offs correspond to high performance but low fairness, we want to guide the search towards regions of higher fairness (by choosing a lower $\alpha$).
Conversely, if our currently explored trade-offs correspond to high fairness but low performance, we want to guide the search towards regions of higher performance (by choosing a higher $\alpha$).

%%%%%%%%%%%%%%%%%%%%%%%%%%%%%%%
%% Dynamic-alpha Explanation %%
%%%%%%%%%%%%%%%%%%%%%%%%%%%%%%%
To achieve the aforementioned balance we need a proxy-metric of our target direction of change.
This direction is given by the difference, $\delta$, between the expected model fairness, $\mathbb{E}_{\lambda \in D}[\phi(\lambda)] = \overline{\phi}$, and the expected model predictive performance, $\mathbb{E}_{\lambda \in D}[\rho(\lambda)] = \overline{\rho}$:
\begin{equation}
\label{eq:delta}
    \delta = \overline{\phi} - \overline{\rho}, \ \delta \in \left[-1, 1\right] .
\end{equation}
Expected values are measured as the mean of respective metric over the sampled hyperparameter configurations, $D  \subseteq \Lambda$.

Hence, when this difference is negative ($\overline{\phi} < \overline{\rho}$), the models we sampled thus far tend towards better-performing but unfairer regions of the hyperparameter space.
Consequently, we want to decrease $\alpha$ to direct our search towards fairer configurations.
Conversely, when this difference is positive ($\overline{\phi} > \overline{\rho}$), we want to direct our search towards better-performing configurations, increasing $\alpha$.
We want this change in $\alpha$ to be proportional to $\delta$ by some constant $k > 0$, such that
\begin{equation}
\label{eq:alpha_sigma}
 \frac{d \alpha}{d \delta} = k, \ k \in \mathbb{R}^{+} ,
\end{equation}
and by integrating this equation we get
\begin{equation}
\label{eq:integrated_alpha}
\alpha = k \cdot \delta + c, \ c \in \mathbb{R} ,
\end{equation}
with $c$ being the constant of integration.
Given that $\delta \in \left[-1, 1\right]$ and $\alpha \in \left[0, 1\right]$, the only feasible values for $k$ and $c$ are $k=0.5$ and $c=0.5$.
Thus, the computation of dynamic-$\alpha$ is given by:
% as $\alpha = 0.5$ corresponds to a balance between performance and fairness metrics,
% we want to increase the value of $\alpha$ up to $\alpha = 1$ when targeting better-performing configurations, and decrease it up to $\alpha = 0$ when targeting fairer configurations.
% Additionally, we want this change in $\alpha$ to be proportional to $\delta$ by some constant $c \in \mathbb{R}^{+}$ (see Equation~\ref{eq:alpha_derivative}).
%
% Hence, after scaling down $\delta$ to the interval $\left[-0.5, 0.5\right]$, the computation of dynamic-$\alpha$ is given as follows by Equation~\ref{eq:dynamic_alpha}.
% Lastly, we apply min-max scaling to $\delta$ to scale it down to the interval $\left[0, 1\right]$ (see Equation ~\ref{eq:min_max_scale}). Overall, for a maximization setting, the computation of the dynamic-$\alpha$ can be described by Equation~\ref{eq:dynamic_alpha}. 

% \begin{equation}
% \label{eq:min_max_scale}
%     \alpha = \frac{(\overline{\phi} - \overline{\rho}) - (-1)}{1 - (-1)}
% \end{equation}

% \begin{equation}
% \label{eq:alpha_derivative}
%     \frac{d \alpha}{d \delta} = c \Rightarrow
%     % \alpha = \int{c \, d\delta} \Rightarrow
%     \alpha = k + c \cdot \delta, \ k \in \mathbb{R}%, c \in \mathbb{R}^{+}
% \end{equation}

\begin{equation}
\label{eq:dynamic_alpha}
    \alpha = 0.5 \cdot (\overline{\phi} - \overline{\rho}) + 0.5 .
\end{equation}

% Figure~\ref{fig:dynamic_alpha} shows a plot of Equation~\ref{eq:dynamic_alpha}.
% Note that to adjust this to a minimization setting, a simple change would be required: $\alpha = 0.5 + (\overline{\rho} - \overline{\phi}) / 2$.

Earlier iterations of bandit-based methods are expected to have lower predictive performance (as these are trained on a lower budget), while later iterations are expected to have higher predictive performance.
By computing new values of $\alpha$ at each Fairband iteration, we promote a dynamic balance between these metrics as the search progresses, predictably giving more importance to performance on earlier iterations but continuously moving importance to fairness as performance increases (a natural side-effect of increasing training budget).
The pseudocode for FB-auto is given by Algorithm~\ref{algo:fairband}.

\begin{algorithm}[t]%H
\caption{Pseudocode for FB-auto}
\label{algo:fairband}

\KwInput~maximum budget per configuration $R$, \\
\hphantom{\KwInput} ratio of budget increase $\eta$ (default $\eta = 3$), \\
\hphantom{\KwInput} metrics trade-off $\alpha$ (default $\alpha = \textit{auto}$)

\begin{algorithmic}%[1]
\State $s_{max} \gets \left\lfloor \log_{\eta}{(R)} \right\rfloor$ \Comment{number of SH brackets}
\For{$s \in \{s_{max}, s_{max} - 1, ..., 0\}$}      \Comment{HB loop}
    \State $n \gets \left\lceil \frac{s_{max} + 1}{s + 1} \cdot \eta^s \right\rceil$, $r \gets R\cdot\eta^{-s}$     \Comment{SH parameters} % \Comment{choice of $n$ \textit{versus} $B/n$ trade-off}
    \State $T \gets $ sample $n$ hyperparameter configs.
    \For{$i \in \left\{ 0, ..., s \right\}$}    \Comment{run SH}
        \State $M \gets $ train configs. $T$ with budget $r_i=\left\lfloor r \eta^{i} \right\rfloor$ 
        % \State $M \gets \text{train\_with\_budget(\lambda, \left\lfloor r \eta^{i} \right\rfloor) : \lambda \in T}$
        \State $P, \Phi \gets $ evaluate perf. and fair. metrics for all $M$
        \If{$\alpha = \textit{auto}$}           \Comment{using dynamic $\alpha$}
            \State $\alpha_i \gets 0.5 \cdot (\sum_j{\Phi_j} / \left|\Phi\right| - \sum_j{P_j} / \left|P\right|) + 0.5 $
        \Else                                                       \Comment{using static $\alpha$}
            \State $\alpha_i \gets \alpha$
        \EndIf

        \State $G \gets \{\alpha_i \cdot P[m_{\bm\lambda}] + (1 - \alpha_i) \cdot \Phi[m_{\bm\lambda}] : m_{\bm\lambda} \in M\}$  %  \Comment{compute goal $g$}
        \State $T \gets $ keep top $k = \left\lfloor \left\lfloor n \cdot \eta^{-i} \right\rfloor / \eta \right\rfloor$ from $T$ ordered by $G$

    \EndFor
\EndFor

\State \Return $\bm\lambda^{*}$, configuration with maximal intermediate goal
\end{algorithmic}
\end{algorithm}

%% ON SELECTION ALPHA
The result of the MOO task is a collection of hyperparameter configurations that represent the fairness-performance trade-off.
One could plot all available choices on the fairness-performance space and manually pick a trade-off, according to whichever business constraints or legislation are in place (see examples of Figure~\ref{fig:AOF_selected}).
For Fairband with static $\alpha$, a target trade-off, $\alpha$, has already been chosen for the method’s search phase, and we once again employ this trade-off for model selection (selection-$\alpha$).
For the FB-auto variant of Fairband, aiming for an automated balance between both metrics, we employ the same strategy for setting $\alpha$ as that used during search.
By doing so, the weight of each metric is pondered by an approximation of their true range instead of blindly applying a pre-determined weight.
For instance, if the distribution of fairness is in range $\phi \in [0, 0.9]$ but that of performance is in range $\rho \in [0, 0.3]$, then a balance would arguably be achieved by weighing performance higher, as each unit increase in performance represents a more significant relative change (this mechanism is achieved by Equation~\ref{eq:dynamic_alpha}).
However, at this stage we can use information from all brackets, as we no longer want to promote exploration of the search space but instead aim for a consistent and stable model selection.
Thus, for FB-auto, the selection-$\alpha$ %\footnote{The value used to weigh the two optimization metrics for model selection.}
is chosen from the average fairness and predictive performance of all sampled configurations.

% %%%%%%%%%%%%%%%%%%%%%%%%%%%%%%%%%%%%%%%%%%%%%%%%%%%%%%%%%%%%%%%%%%%%%%%%%%%%%%%%%%%%%%%%%%%
% % 
% %                                 EXPERIMENTAL SETUP
% % 
% % %%%%%%%%%%%%%%%%%%%%%%%%%%%%%%%%%%%%%%%%%%%%%%%%%%%%%%%%%%%%%%%%%%%%%%%%%%%%%%%%%%%%%%%%%

\section{Experimental Setup}

In order to validate our proposal, we evaluate fairness-aware HO on a search space spanning multiple ML algorithms, model hyperparameters, and sampling hyperparameters, on a private large-scale case-study on bank account opening fraud dataset.
For reproducibility, we further evaluate our method on three benchmark datasets from the fairness literature: COMPAS~\cite{Angwin2016}, Adult~\cite{UCI}, and Donors Choose (DC)~\cite{donorschoose2014}.
Details and results for literature datasets are in Appendix~\ref{app:supp_materials}\footnote{Data %, plots, 
and ML artifacts from all datasets at \href{https://github.com/feedzai/fair-automl}{github.com/feedzai/fair-automl}
}.

In our evaluation suite, we include a selection of state-of-the-art HO algorithms (RS, TPE, and HB), together with our proposed fairness-aware versions (FairRS, FairTPE, and Fairband).
%
% We compare our method to several baselines in the HO community, including Random Search (RS) and Hyperband (HB).
For fairness-aware algorithms, we employ $\alpha=0.5$. FB-auto uses our proposed dynamic-$\alpha$ heuristic.
% We study two versions of Fairband: FB-auto (employing the dynamic $\alpha$ strategy) and FB (employing $\alpha = 0.5$).

\subsection{Account Opening Fraud Dataset}

In order to validate our proposal, we evaluate fairness-aware HO on a large-scale case-study (dubbed \textbf{AOF}) from real-world bank account opening applications, spanning a 6-month data stream with over 500K instances.
In account opening fraud, a malicious actor will attempt to open a new bank account using a stolen or synthetic identity (or both), in order to quickly max out its line of credit~\cite{FraudSurvey2016}.
Banks have to comply with anti-fraud regulations, and are liable for incurred losses when fraudsters gain illicit access to a credit line.
At the same time, 
holding a bank account has been deemed a basic right by the European Union, region where our case-study takes place~\cite{basic_account_EU}.

Our objective is to maximize fraud recall while maintaining an FPR below 5\% (a business constraint to minimize customer churn).
Fairness-aware methods have the additional objective of equalizing FPR across age groups (preventing ageism), also known as predictive equality~\cite{Corbett-Davies2017}.
This is formalized as maximizing the ratio between the smallest group-wise FPR and the largest group-wise FPR~\cite{Saleiro2018}:
\begin{equation}
    \frac{\min_{a \in A}{P \left[ \hat{Y}=1 | A=a, Y=0 \right]}}{\max_{a \in A}{P \left[ \hat{Y}=1 | A=a, Y=0 \right]}} ,
\end{equation}
where $A$ is the set of sensitive attributes (in our case, age groups).
%
% Thus our goal is to prevent ageism while maximizing fraud recall.
%
Note that a false positive is a legitimate customer that sees her/his request unjustly denied, disturbing the customer's life and costing the bank potential earnings.
% Moreover, by optimizing for predictive equality we place the burden of a low-performing model squarely on the financial institution that is responsible for it, not on the customers that have no power over the model selection process.

%% SEARCH SPACE %%
\subsection{Search Space}

Firstly, we consider a broad hyperparameter space as a requirement for the effective execution of fairness-aware HO.
We consider as hyperparameters any decision in the ML pipeline, as bias can be introduced at any stage of this pipeline~\cite{barocas2016big}.
Both performance and fairness metrics are seen as (possibly noisy) black-box functions of these hyperparameters.
Thus, an effective search space includes which model type to use, the model hyperparameters which dictate how it is trained, and the sampling hyperparameters which dictate the distribution and prevalence rates of training data.

In order to validate our methods, we define a comprehensive hyperparameter search space, allowing us to effectively navigate the fairness-performance trade-off.
We select five ML model types: LightGBM (LGBM)~\cite{LGBM}, Random Forest (RF)~\cite{Breiman2001}, Decision Tree (DT)~\cite{breiman1984classification}, Logistic Regression (LR)~\cite{Walker1967}, and feed-forward neural network (NN).
Sampling a new hyperparameter configuration is seen as a hierarchical process, where we first select the model type, and afterwards we sample its model-specific hyperparameters.
%
% Besides the model type and model hyperparameters, we also explore three different undersampling strategies: targeting 20\%, 10\%, and 5\% positive samples.
% This type of hyperparameter is only used on the AOF dataset, to tackle its high class imbalance\footnote{The AOF dataset features approximately 99 negatively-labeled samples per each positively-labeled sample, as most bank account applicants are genuine.}.
%
In order to tackle the severe class imbalance on the AOF dataset\footnote{The AOF dataset features approximately 99 negatively-labeled samples per each positively-labeled sample, as most bank account applicants are genuine.}, we further explore three undersampling strategies: targeting 20\%, 10\%, and 5\% positive samples.

% \subsection{Hyperband Parameters}
% \subsection{HO Parameters}
\subsection{HO Parameters}
\label{sec:hyperband_params}

For resource-aware HO algorithms, we define 1 budget unit as 1\% of the training dataset.
The maximum budget is defined as $R=100$ (100\% of the training dataset), and HB's budget increase ratio is defined as $\eta=3$.
With these parameters, HB will run $s_{max} + 1 = \left\lfloor \log_{\eta}{(R)} \right\rfloor + 1 = 5$ iterations of SH.
%
% Table~\ref{tab:hyperband_brackets} displays the number of sampled hyperparameter configurations %per SH bracket and 
% per iteration, together with the budget per configuration.
%
Each bracket will use at most $s_{max} + 1 = 5$ training slices of increasing size, corresponding to the following dataset percentages: 1.23\%, 3.70\%, 11.1\%, 33.3\%, 100\%.

% The outer loop will run $s_{max} + 1 = 5$ times, for $s \in \{s_{max}, ..., 0\}$.
% Each run (or bracket) will consume at most $B=500$ budget units.
%
%
The training slices are sampled such that smaller slices are contained in larger slices, and such that the class-ratio is maintained (by stratified sampling).
The number of configurations and budget per configuration within each bracket are displayed in Table~\ref{tab:hyperband_brackets}.
%
% Throughout a complete Hyperband run, 143 unique hyperparameter configurations will be randomly sampled (total $n_i$ for $i=0$), and 206 models will be trained and evaluated (total $n_i$ for $i \in \{0 ..., s_{max}\}$).

% %%%%%%%%%%%%%%%%%%%%%%%%%%%%%%%%%%%%%%%%%%%%%%%%%%%%%%%%%%%%%%%%%%%%%%%%%%%%%%%%%%%%%%%%%%%%%%%%%%%%%%%%%%%%%%%%
% % 
% %                                               RESULTS
% % 
% % %%%%%%%%%%%%%%%%%%%%%%%%%%%%%%%%%%%%%%%%%%%%%%%%%%%%%%%%%%%%%%%%%%%%%%%%%%%%%%%%%%%%%%%%%%%%%%%%%%%%%%%%%%%%%
\section{Results \& Discussion}

In this section, we present and analyze the results from our fairness-aware HO experiments on the AOF case-study.
Results for literature datasets are in Appendix~\ref{app:supp_materials}.
To validate the methodology, we guide hyperparameter search by evaluating all sampled configurations on the same validation dataset, while evaluating the best-performing configuration (that maximizes Equation~\ref{eq:goal}) on a held-out test dataset at the end.
Likewise, the model thresholds are set on the validation dataset and then used on both the validation and test datasets, mimicking a production environment.
All studied HO methods are given the same training budget: 2400 budget units.
% For RS, this corresponds to training 24 distinct configurations on 100\% of the training dataset.
% For HB and FB, it corresponds to a full run with $\eta=3$ and $R=100$.

\begin{table}[tb]
    \caption{
    Validation and test results on the AOF dataset.
    % Statistical significance for fairness-aware methods is tested against their respective fairness-blind counterparts with a Kolmogorov-Smirnov test~\cite{doi:10.1080/01621459.1967.10482916}.
    % Results are marked with a $\lozenge$ when $p < 0.05$, and a $\blacklozenge$ when $p < 0.01$.
    }
    \centering
    \input{tables/AOF_results_15_runs}
    \label{tab:AOF_results_table}
\end{table}

\begin{figure*}[t]
\begin{subfigure}[b]{0.49\textwidth}
    \centering
    \includegraphics[width=1.0\columnwidth]{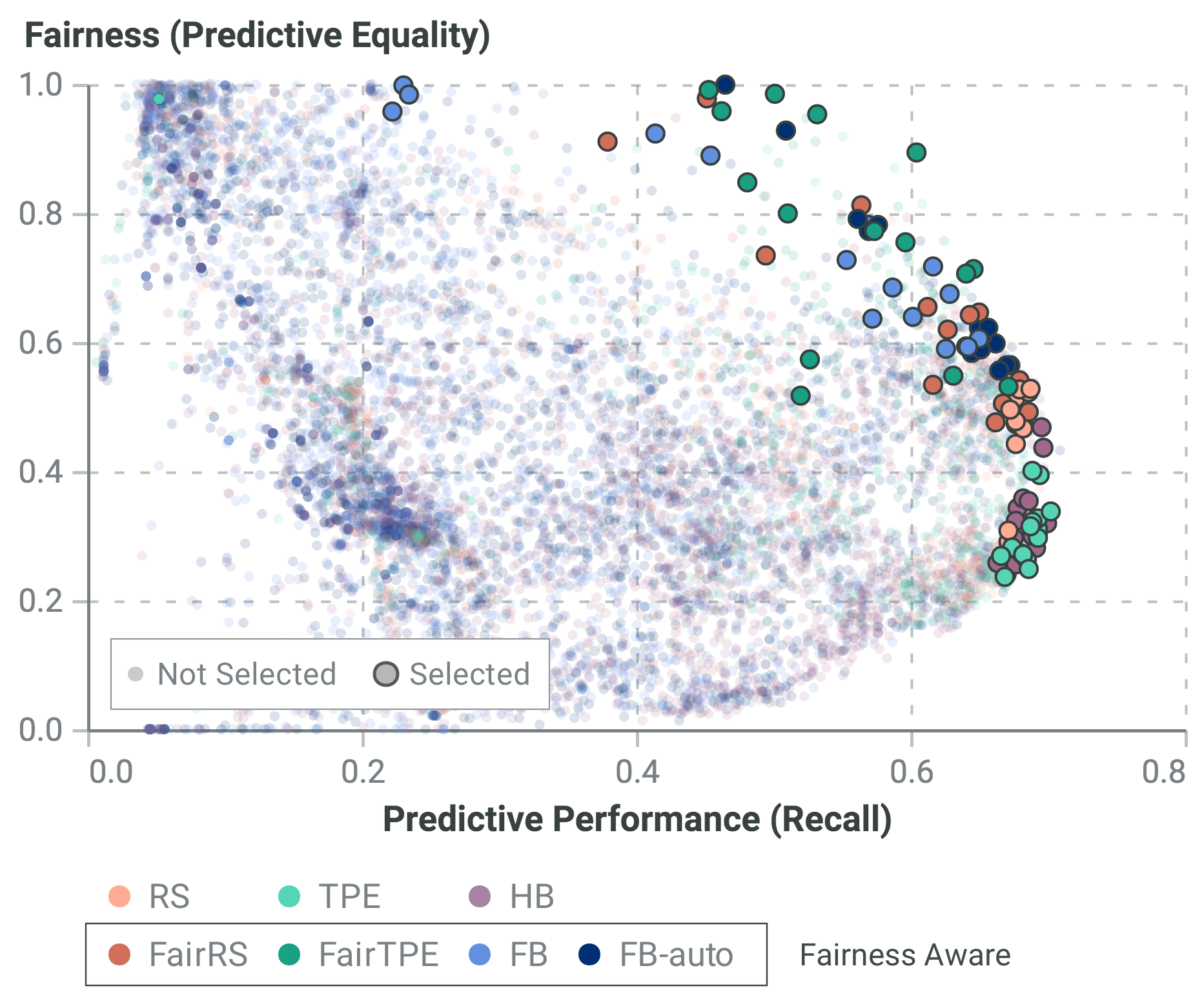}
    \caption{}
    \label{fig:AOF_selected}
\end{subfigure}
\hfill
\begin{subfigure}[b]{0.49\textwidth}
    \centering
    \includegraphics[width=1.0\columnwidth]{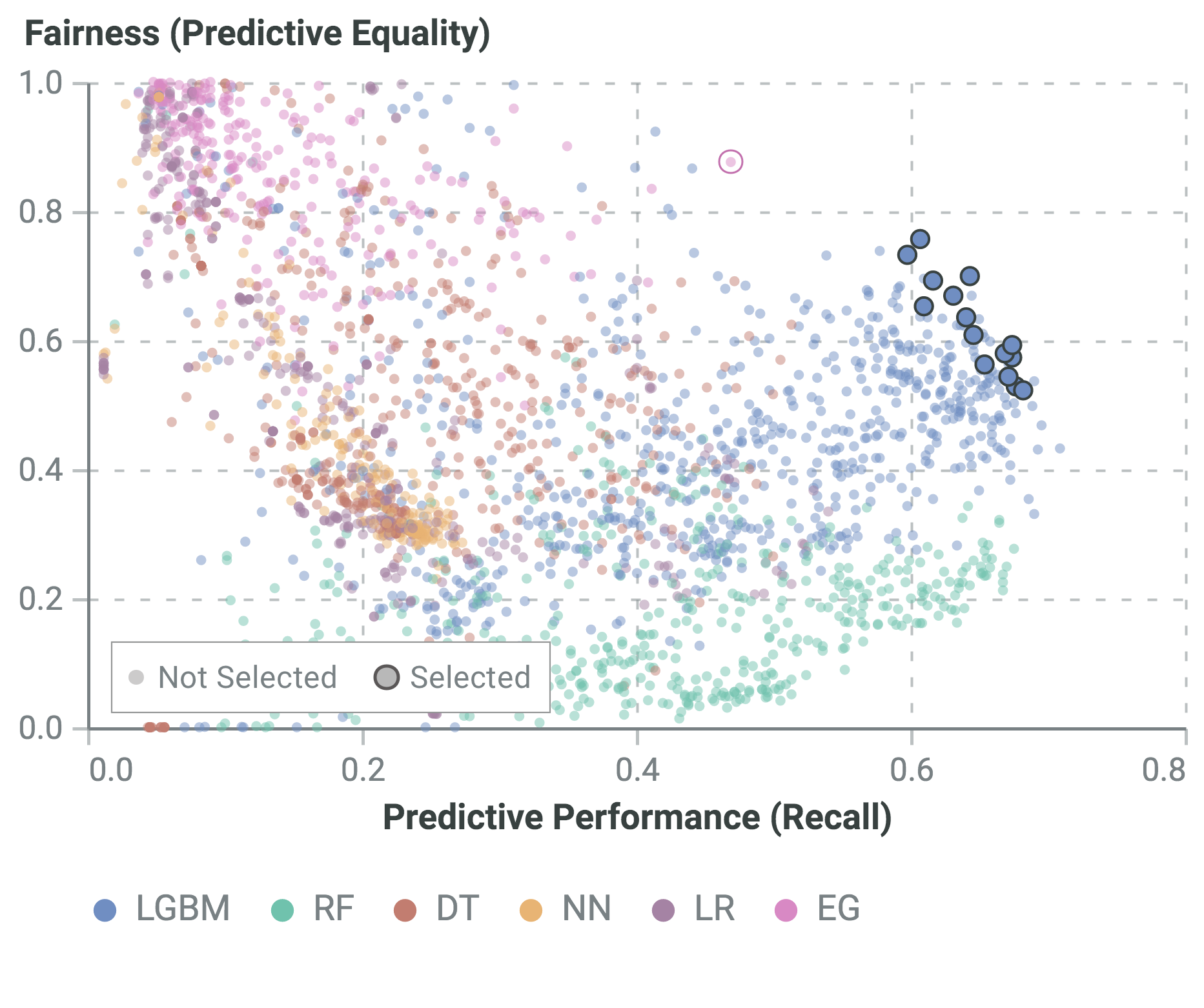}
    \caption{}
    \label{fig:AOF_fairlearn}
\end{subfigure}

\caption{
    Fairness and performance of models sampled (transparent smaller circles) and selected (larger circles) by evaluated hyperparameter tuners, on the AOF dataset (validation data).
    Figure~\ref{fig:AOF_selected} (left) corresponds to the results of all tuners on Table~\ref{tab:AOF_results_table}.
    Figure~\ref{fig:AOF_fairlearn} (right) corresponds to the results using only FB-auto, and after adding EG~\cite{Agarwal2018} to the search space (Section~\ref{sec:AOF_fairlearn}).
    % Figure~\ref{fig:AOF_fairlearn} (right) corresponds to the experiment described in Section~\ref{sec:AOF_fairlearn}, using only the FB-auto tuner, and after adding the EG in-processing bias reduction method~\cite{Agarwal2018} to the search space.
}
\label{fig:AOF}
\end{figure*}

% \subsection{General Results} 
Table~\ref{tab:AOF_results_table} shows the validation and test results of running traditional HO methods on the AOF dataset, together with our proposed fairness-aware variants.
Results are averaged over 15 runs, and 
% statistical significant differences against the baseline methods (fairness-blind counterparts) are shown with $\blacklozenge$.
statistical significance for fairness-aware methods is tested against their respective fairness-blind counterparts (\eg, FB vs HB) with a Kolmogorov-Smirnov test~\cite{doi:10.1080/01621459.1967.10482916} ($\lozenge$ when $p < 0.05$, $\blacklozenge$ when $p < 0.01$).
This table comprises the training and evaluation of over 10K unique models. Together with the results on literature datasets we total over 40K models, one of the largest studies of the fairness-performance trade-off yet.
%
% Our methods consistently achieve higher fairness than the baselines on all datasets, at a small cost in predictive performance (statistically significant for all tuners on validation data).

The differences in predictive performance and fairness between the proposed fairness-aware methods and their fairness-blind counterparts have strong statistical significance for all proposed HO algorithms on validation data.
This is visible on test data as well, except for FairRS, which achieves significant fairness improvements with statistically insignificant performance decreases.
This clear overall trend is also visible on the literature datasets.
Additionally, FB-auto shows higher predictive performance than both FB and FairTPE ($p < 0.05$), while differences in fairness are not statistically significant for these pairs.
In fact, the average fairness-performance trade-offs are Pareto optimal for all tuners except FairRS when measured among themselves, each representing a distinct trade-off.

Overall, FB-auto and FairTPE arguably achieve the best fairness-performance trade-offs to be used in a production environment.
FB-auto achieves an averaged 111\% improvement on the fairness metric, accompanied by a comparatively small 6\% decrease on the performance metric, when compared with HB.
FairTPE achieves a 137\% fairness improvement, accompanied by a 14\% decrease in performance, when compared with TPE.
With these results we emphasize that, due to the empirical Pareto frontier geometry, fairness increases are not proportional to performance decreases: steep fairness increases can be achieved with small performance decreases.
From a practical standpoint, models selected by FB-auto on average approve 411 more bank accounts per month from legitimate older-aged individuals that would have been rejected by the HB-selected model. % (for the same global FPR).
% FairTPE models approve 522 more accounts per month from this group when compared with vanilla TPE.

%% FB-auto vs HB: 411 more
%% FB vs HB: 510 more
%% FairRS vs RS: 247 more
%% FairTPE vs TPE: 522 more

\begin{figure}[tb]
    \centering
    \includegraphics[width=0.6\columnwidth]{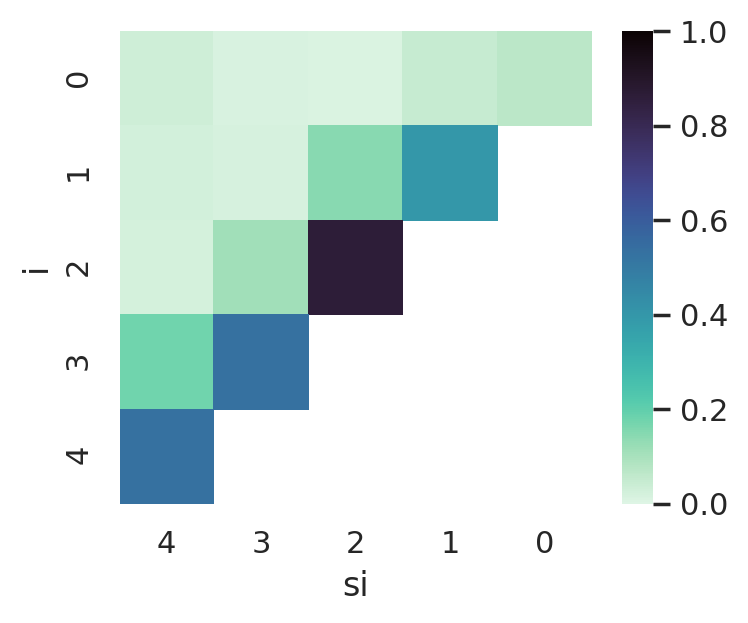}
    \caption{Average density of Pareto optimal models per FB-auto iteration on the AOF dataset (iteration details on Table~\ref{tab:hyperband_brackets}). % for information on each iteration.
    }
    \label{fig:AOF_heatmap}
\end{figure}

% Search Strategy
\subsection{Search Strategy} 
\label{sssec:search_strategy}

We evaluate the Fairband search strategy by analyzing the evolution of fairness and performance simultaneously, and whether we can effectively extend the practical Pareto Frontier as the search progresses.
That is, whether optimal trade-offs are more likely to be found as we discard the worst performing models and further increase the allocated budget for top-performing models within each bracket (see Table~\ref{tab:hyperband_brackets}).
We focus on FB-auto, the fairness-aware variant of HB that automatically tunes the fairness-performance trade-off.
%
%In order to evaluate our method's search strategy, we must study the evolution of fairness and performance simultaneously, and whether we can effectively extend the practical Pareto frontier as the search progresses.
%
% Figure~\ref{fig:heatmaps} shows a heat map of the density of Pareto optimal models\footnote{Fraction of Pareto optimal models within each bracket, with optimality assessed within each run.} in each FB iteration, for FB-auto, on the datasets Adult and AOF
% \footnote{Similar results are observed on the other datasets.}.
Figure~\ref{fig:AOF_heatmap} shows a heat map of the average density of Pareto optimal models
% \footnote{Average fraction of Pareto optimal models within each bracket.%, averaged over 15 runs.}
in each model-discard iteration, for 15 runs of the FB-auto algorithm, on the AOF dataset. %(similar results are observed on all datasets).
As the iterations progress, under-performing configurations are pruned, and the density of Pareto optimal models steadily increases, confirming the effectiveness of the search strategy.

%%%% -----------------------------------
%%%         MODEL SELECTION 
%%%% -----------------------------------
\subsection{Model Selection}

Figure~\ref{fig:AOF_selected} shows the final selected models for 15 runs of each method on the AOF dataset.
The remaining models considered during the search are also shown, with lower opacity and smaller size, and with equal coloring pattern as selected models.
Fairband (blues) consistently identifies good fairness-performance trade-offs from the universe of available configurations.
Indeed, the models selected by Fairband are consistently close to or form the Pareto frontier.
On the other hand, FairTPE seems to be able to find the best fairness-performance trade-offs, but with less reliable results.
Crucially, as evident by the spread of selected models, we can successfully navigate the fairness-performance Pareto frontier solely by means of HO.
It is important to consider that the studied fairness-blind methods are the current standard in HO.
By unfolding the fairness dimension, we show that strong predictive performance carries an equally strong real-world cost in unfairness, which is hidden by traditional HO methods.

%%%%%%%%%%%%%%%%%%%%%%%%%%%%%%%%%%%%%%%
% FAIRLEARN EXPERIMENT
%%%%%%%%%%%%%%%%%%%%%%%%%%%%%%%%%%%%%%%

\subsection{Optimizing Bias Reduction Hyperparameters}
\label{sec:AOF_fairlearn}

As an HO method, one fruitful approach to bias-mitigation is including traditional bias reduction methods into our hyperparameter search space.
We introduce the Exponentiated Gradient (EG) reduction for fair classification algorithm~\cite{Agarwal2018} into our search space on the AOF dataset.
EG is a state-of-the-art bias reduction algorithm that optimizes predictive performance subject to fairness constraints, and is compatible with any cost-sensitive binary classifier.
In our setting, we target predictive equality (the fairness metric on AOF), and apply EG over a Decision Tree classifier.

Figure~\ref{fig:AOF_fairlearn} shows a plot of the models selected by FB-auto over 15 runs on the AOF dataset, discriminated by model type.
As can be seen, LGBM models on this particular task generally dominate all others.
Only a single model on the Pareto frontier is not LGBM, and is indeed EG (circled in pink).
However, all models selected by FB-auto are LGBMs, and these arguably represent the best fairness-performance trade-offs.
We note that different model types occupy distinct regions of the fairness-performance space (visible both in Figure~\ref{fig:AOF_fairlearn} and appendix Figure~\ref{fig:adult_fairlearn}), 
hinting that fairness may not be independent of model type.
% These results also suggest that fairness is sensitive to a multitude of hyperparameters, including the model type, as is the case of performance.
%
While including bias reduction methods in the HO search space may be useful for extending the Pareto frontier on some tasks, using \textit{only} bias reduction methods may lead to severely worse fairness-performance trade-offs.
% Overall, 
These results support the fact that fairness-aware HO should be employed in all ML pipelines that aim for fair decision-making, together with a wide selection of ML algorithms to properly explore the fairness-performance search-space.

% %%%%%%%%%%%%%%%%%%%%%%%%%%%%%%%%%%%%%%%%%%%%%%%%%%%%%%%%%%%%%%%%%%%%%%%%%%%%%%%%%%%%%%%%%%%%%%%%%%%%%%%%%%%%%%%%
% % 
% %                                               CONCLUSION
% % 
% % %%%%%%%%%%%%%%%%%%%%%%%%%%%%%%%%%%%%%%%%%%%%%%%%%%%%%%%%%%%%%%%%%%%%%%%%%%%%%%%%%%%%%%%%%%%%%%%%%%%%%%%%%%%%%
 \section{Conclusion}
 \label{sec:conclusions}

There have been widespread reports of real-world ML systems shown to be biased, causing serious disparate impact across different sub-groups, unfairly affecting people based on race, gender or age. Although the ML research community has embraced this issue, the current landscape of algorithmic fairness still lacks (1) practical methodologies and (2) tools for real-world practitioners.

This work aims to bridge that gap by providing a simple and flexible intervention to foster the incorporation of fairness objectives in real-world ML pipelines: fairness-aware HO.
We propose and evaluate fairness-aware variants of state-of-the-art HO algorithms: Fair Random Search, Fair TPE, and Fairband.
%
% These fairness-aware methods jointly optimize predictive performance and fairness metrics by guiding the hyperparameter search towards fairer configurations. %navigating their hyperparameter search space.
%
We further propose a heuristic for setting the relative fairness-performance weight on a bandit-based context (FB-auto).
% Fairband-auto implements this heuristic and 

By introducing fairness notions into HO, our method can be seamlessly integrated into real-world ML pipelines, at no extra training cost.
Moreover, our method is easy to implement, resource-efficient, and both model- and metric-agnostic, providing no obstacles to its widespread adoption.

We evaluate our method on a large-scale online account opening fraud case-study as well as on three benchmark datasets from the fairness literature.
Fairness-aware HO is shown to provide significant fairness improvements at a small cost in predictive performance, when compared to traditional HO techniques.
On the AOF dataset, FB-auto achieves 111\% averaged improvement on the fairness metric, accompanied by a comparatively small 6\% decrease on the performance metric, when compared with HB.

We show that it is both possible and effective to navigate the fairness-performance trade-off through HO.
At the same time, we observe that there is a wide spread of attainable fairness values at any level of predictive performance, and once again document the known inverse relation between fairness and predictive performance.
Crucially, we empirically show that by only optimizing for a predictive performance metric (as is standard practice in real-world ML systems) we unknowingly target unfairer regions of hyperparameter space.

\section*{Acknowledgment}
% The project CAMELOT (reference POCI-01-0247-FEDER-045915) leading to this work is co-financed by the ERDF - European Regional Development Fund through the Operational Program for Competitiveness and Internationalisation - COMPETE 2020, the North Portugal Regional Operational Program - NORTE 2020 and by the Portuguese Foundation for Science and Technology - FCT under the CMU Portugal international partnership.
%
We would like to express our gratitude to Beatriz Malveiro for invaluable feedback on the paper's visualizations.

\bibliographystyle{unsrt}
\bibliography{refs}

\vspace*{65em}

% \pagebreak
% \clearpage

\begin{table*}[tb]
\caption{Classification task details for each literature dataset.}
\centering
\begin{tabular}{lcccccc}
\toprule
\textbf{Dataset} & \textbf{Setting} & \textbf{Pred. Acc. Metric} & \textbf{Fairness Metric} & \textbf{Target Threshold} & \textbf{Sensitive Attr.} & \textbf{Prediction Task} \\
\midrule
DC & assistive & precision & equal opportunity & 1000 PP & school poverty level & risk of underfunding project \\
Adult   & assistive & precision & equal opportunity & 50\% TPR & gender & income $\leq$ \$50K/year \\
COMPAS  & punitive  & precision & predictive equality & 10\% FPR & race & recidivism risk \\
% AOF     & punitive & recall & predictive equality & 5\% FPR & age & account opening fraud \\
\bottomrule
\end{tabular}
\label{tab:literature_datasets}
\end{table*}

\appendix%[Supplementary Materials]
\label{app:supp_materials}

\subsection{Experiments on Public Datasets}

We further validate our methodology on three benchmark datasets from the fairness literature. %: COMPAS~\cite{Angwin2016}, Adult~\cite{UCI}, and Donors Choose~\cite{donorschoose2014}.
%
% Aiming to cover a wide range of ML tasks, we elaborate a distinct context for each dataset.
% The choice of performance and fairness metrics is highly dependent on the task for which a given model is trained, and the real-world setting in which it will be deployed.
%
% We choose performance and fairness metrics that fit the real-world use-case of each dataset~\cite{bigdatasocialsciences}. %, as well as target prediction thresholds.
%
We use equal opportunity~\cite{Hardt2016} (\ie, balanced true positive rates) for assistive tasks% (\ie, a positive prediction is favorable to the individual), 
, and we employ predictive equality~\cite{Corbett-Davies2017} (\ie, balanced FPRs) for punitive tasks% (\ie, a positive prediction is detrimental to the individual)
~\cite{Saleiro2018}.
Table~\ref{tab:literature_datasets} summarizes the classification task for each dataset.

The \textbf{Donors Choose} (DC) dataset~\cite{donorschoose2014} consists of data pertaining to thousands of projects proposed for/by K-12 schools.
The objective is to identify projects at risk of getting underfunded to provide tailored interventions.
We set a limit of 1000 positive predictions (PP) as a realistic budget for this task.
The \textbf{Adult} dataset~\cite{UCI} consists of data from the 1994 US census, including age, gender, race, occupation, and income, among others.
We devise a hypothetical assistive program that targets individuals making less than \$50K/year (opposite of the original label of ``$>$\$50K/year'').
%\footnote{This is the opposite of the standard target label ``$>$\$50K/year''.}.
%
%% SIZE: 6172
The \textbf{COMPAS} dataset~\cite{Angwin2016} is a real-world criminal justice dataset.
The objective is predicting whether someone will re-offend based on the person's criminal history, demographics, and prior jail time.

\begin{table}[tbh]
    \caption{
    Validation and test results on the DC, Adult, and COMPAS datasets ($\lozenge$ when $p < 0.05$, $\blacklozenge$ when $p < 0.01$).
    }
    \centering
    \input{tables/literature_results}
    \label{tab:literature_results}
\end{table}

Table~\ref{tab:literature_results} displays results from all evaluated methods on benchmark datasets from the literature.
%
% Figures~\ref{fig:Adult_selected},~\ref{fig:COMPAS_selected}, and \ref{fig:DC_selected} show selected models for Adult, COMPAS, and Donors Choose datasets, respectively.
%
We note that the Adult and Donors Choose datasets are approximately one order of magnitude smaller than AOF, while COMPAS is two orders of magnitude smaller.
Overall results are in accordance with those of the AOF case-study.

\subsection{Optimizing Bias Reduction Hyperparams. on Adult Dataset}

\begin{figure}[tbh]
    \centering
    \includegraphics[width=1.0\columnwidth]{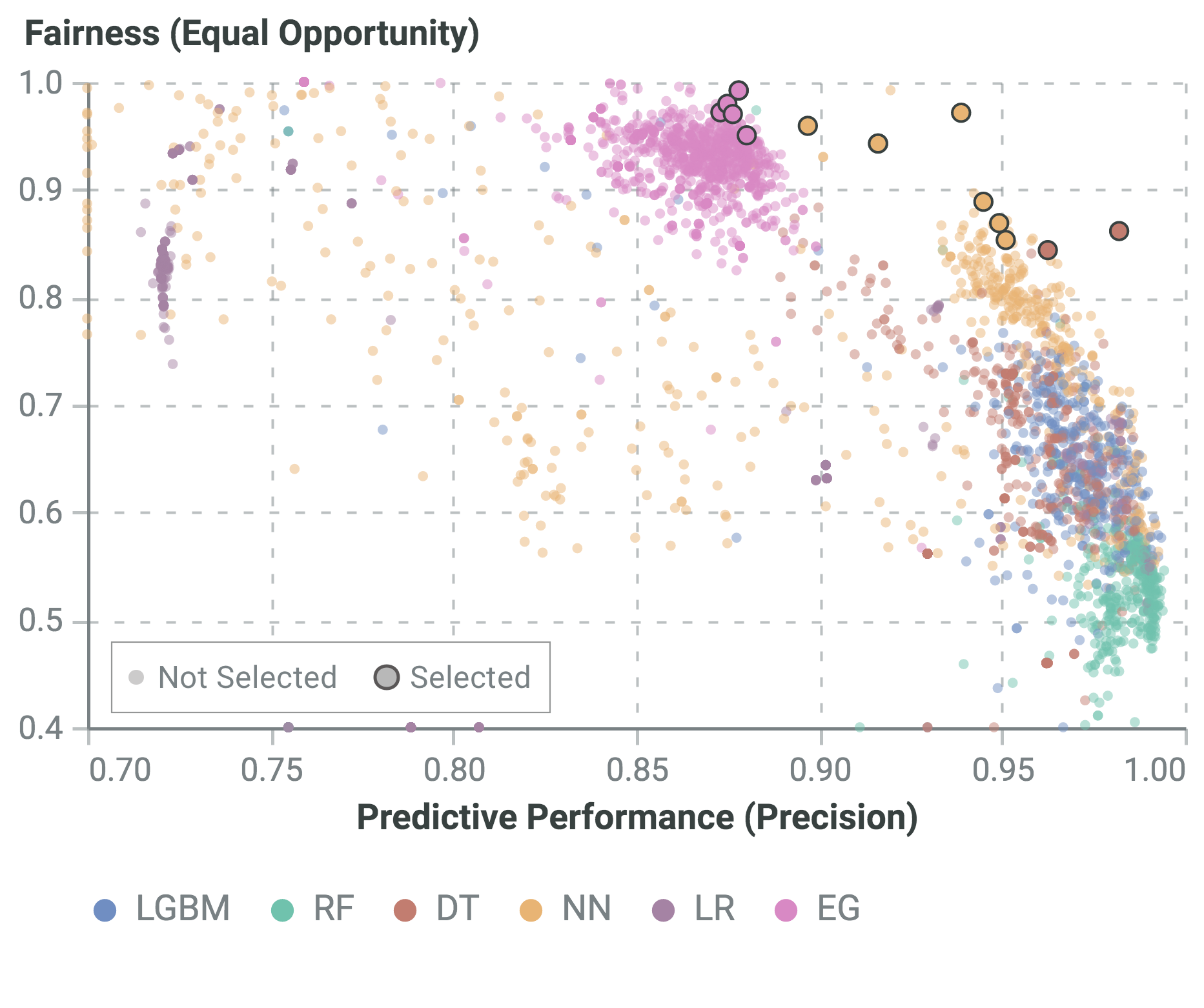}
    % \includegraphics[width=0.8\columnwidth]{figures/plots/Adult-fairlearn.png}
    % \vspace{-3mm}
    \caption{Fairness and predictive performance of models sampled (smaller circles) and selected (larger circles) by FB-auto, discriminated by model type, on the Adult dataset. %Search space includes the in-processing bias reduction method \textit{Exponentiated Gradient} (EG) reduction~\cite{Agarwal2018}, used together with Decision Tree classifiers.
    }
    \label{fig:adult_fairlearn}
\end{figure}

Figure~\ref{fig:adult_fairlearn} shows a plot of the models selected by FB-auto over 15 runs on the Adult dataset.
The introduction of EG~\cite{Agarwal2018} creates a new cluster of models in our search space (shown in pink), consisting of possible fairness-performance trade-offs in a previously unoccupied region.
% (compare with Figure~\ref{fig:Adult_selected}).
%
However, even though these models were trained specifically targeting our fairness metric while the remaining models were trained in a fairness-blind manner, FB-auto chooses other model types more often than not.
Indeed, the selected NNs and DTs arguably represent the best fairness-performance trade-offs.
Interestingly, the types of ML models on the Pareto frontier are markedly different from those observed on the AOF dataset (Figure~\ref{fig:AOF_fairlearn}).
However, once again, RF consistently displays worse fairness-performance trade-offs than LGBM. %, and these relations between specific hyperparameters and better/worse fairness will be studied in future work.
Similarly to the results on the AOF dataset, it is shown that blindly applying bias reduction techniques may lead to sub-optimal fairness-performance trade-offs.

\end{document}

%% file: tables/hyperband_brackets.tex
\begin{tabular}{c|cc|cc|cc|cc|cc}
% \fontsize{9}{11}\selectfont
% \toprule
& \multicolumn{2}{c|}{$s = 4$} & \multicolumn{2}{c|}{$s = 3$} & \multicolumn{2}{c|}{$s = 2$} & \multicolumn{2}{c|}{$s = 1$} & \multicolumn{2}{c}{$s = 0$} \\
$i$	& $n_i$ & $r_i$		& $n_i$ & $r_i$		& $n_i$ & $r_i$ & $n_i$ & $r_i$ & $n_i$ & $r_i$ \\ \midrule
0	& 81    & 1.2		& 34    & 3.7		& 15 & 11     & 8 & 33	& 5 & 100  \\
1	& 27    & 3.7		& 11    & 11		& 5  & 33     & 2 & 100	&   &   \\
2	& 9     & 11		& 3     & 33		& 1  & 100      &   & 		&   &   \\
3	& 3     & 33		& 1     & 100		&	 &          &   & 		&   &   \\
4	& 1     & 100		&       & 			&	 &          &   & 		&   &   \\
% \bottomrule
\end{tabular}

%% file: tables/AOF_results_15_runs.tex
\begin{tabular}{lcccc}

\toprule
     \multirow{2}{*}{\textbf{Algo.}} & \multicolumn{2}{c}{\textbf{Validation}}       & \multicolumn{2}{c}{\textbf{Test}}            \\
     & \textbf{Pred. Perf.} & \textbf{Fairness} & \textbf{Pred. Perf.} & \textbf{Fairness} \\
\midrule

FB-auto &   $ 61.7^\blacklozenge $ &  $ 68.1^\blacklozenge $ &  $ 64.0^\blacklozenge $ &  $ 74.2^\blacklozenge $ \\
    FB &   $ 50.7^\blacklozenge $ &  $ 76.0^\blacklozenge $ &  $ 52.6^\blacklozenge $ &  $ 81.3^\blacklozenge $ \\
FairRS &   $ 60.4^\blacklozenge $ &  $ 64.1^\blacklozenge $ &  $ 62.6 $ &  $ 68.6^\blacklozenge $ \\
FairTPE &   $ 55.7^\blacklozenge $ &  $ 76.9^\blacklozenge $ &  $ 59.2^\blacklozenge $ &  $ 80.0^\blacklozenge $ \\
    HB &    $ 68.4 $ &  $ 32.3 $ &  $ 68.4 $ &  $ 35.2 $ \\
    RS &    $ 67.8 $ &  $ 42.2 $ &  $ 68.1 $ &  $ 45.0 $ \\
   TPE &    $ 68.7 $ &  $ 30.5 $ &  $ 68.5 $ &  $ 33.7 $ \\
\bottomrule
\end{tabular}

%% file: tables/literature_results.tex
\begin{tabular}{lcccc}

\toprule
     \multirow{2}{*}{\textbf{Algo.}} & \multicolumn{2}{c}{\textbf{Validation}}       & \multicolumn{2}{c}{\textbf{Test}}            \\
     & \textbf{Pred. Perf.} & \textbf{Fairness} & \textbf{Pred. Perf.} & \textbf{Fairness} \\
\midrule
%^{\lozenge\blacklozenge}
  \multicolumn{5}{c}{Donors Choose} \\
\midrule
FB-auto &   $ 54.2^\blacklozenge $ &  $ 98.2^\blacklozenge $ &  $ 50.7^\blacklozenge $ &  $ 86.5^\blacklozenge $ \\
 FB &   $ 54.2^\blacklozenge $ &  $ 97.7^\blacklozenge $ &  $ 50.4^\blacklozenge $ &  $ 85.5^\blacklozenge $ \\
 FairRS &   $ 51.7^\blacklozenge $ &  $ 97.0^\blacklozenge $ &  $ 50.4^\blacklozenge $ &  $ 79.5^\blacklozenge $ \\
FairTPE &   $ 52.3^\blacklozenge $ &  $ 96.3^\blacklozenge $ &  $ 50.6^\blacklozenge $ &  $ 79.1^\blacklozenge $ \\
    HB &    $ 60.9 $ &  $ 28.7 $ &  $ 53.6 $ &  $ 35.0 $ \\
    RS &    $ 59.9 $ &  $ 24.9 $ &  $ 53.4 $ &  $ 32.4 $ \\
   TPE &    $ 61.0 $ &  $ 27.1 $ &  $ 53.3 $ &  $ 33.4 $ \\
\midrule

  \multicolumn{5}{c}{Adult} \\
\midrule
FB-auto &   $ 92.0^\blacklozenge $ &  $ 94.7^\blacklozenge $ &  $ 91.6^\blacklozenge $ &  $ 90.9^\blacklozenge $ \\
 FB &   $ 92.7^\blacklozenge $ &  $ 94.0^\blacklozenge $ &  $ 92.3^\blacklozenge $ &  $ 89.5^\blacklozenge $ \\
 FairRS &   $ 93.6^\blacklozenge $ &  $ 79.4^\blacklozenge $ &  $ 93.8^\blacklozenge $ &  $ 78.6^\blacklozenge $ \\
FairTPE &   $ 93.3^\blacklozenge $ &  $ 82.2^\blacklozenge $ &  $ 93.5^\blacklozenge $ &  $ 80.7^\blacklozenge $ \\
    HB &    $ 99.4 $ &  $ 53.5 $ &  $ 99.0 $ &  $ 54.1 $ \\
    RS &    $ 99.4 $ &  $ 55.7 $ &  $ 99.1 $ &  $ 56.6 $ \\
   TPE &    $ 99.4 $ &  $ 54.9 $ &  $ 99.1 $ &  $ 55.6 $ \\
\midrule

  \multicolumn{5}{c}{COMPAS} \\
\midrule
FB-auto &   $ 74.0^\blacklozenge $ &  $ 95.8^\blacklozenge $ &  $ 70.1^\lozenge $ &  $ 90.0^\blacklozenge $ \\
 FB &   $ 71.2^\blacklozenge $ &  $ 95.5^\blacklozenge $ &  $ 67.6^\blacklozenge $ &  $ 80.7^\blacklozenge $ \\
 FairRS &   $ 67.4^\blacklozenge $ &  $ 77.4^\blacklozenge $ &  $ 64.2^\blacklozenge $ &  $ 67.8^\blacklozenge $ \\
FairTPE &   $ 67.1^\blacklozenge $ &  $ 81.8^\blacklozenge $ &  $ 63.9^\blacklozenge $ &  $ 69.5^\blacklozenge $ \\
    HB &    $ 78.1 $ &  $ 45.4 $ &  $ 73.6 $ &  $ 51.2 $ \\
    RS &    $ 77.7 $ &  $ 43.8 $ &  $ 73.2 $ &  $ 43.4 $ \\
   TPE &    $ 78.0 $ &  $ 42.8 $ &  $ 73.5 $ &  $ 46.6 $ \\
% \midrule

%   \multicolumn{5}{c}{AOF} \\
% \midrule
% FB-auto &   $ 61.7 $ &  $ 68.1 $ &  $ 64.0 $ &  $ 74.2 $ \\
%  FB &   $ 50.7 $ &  $ 76.0 $ &  $ 52.6 $ &  $ 81.3 $ \\
%  FairRS &   $ 60.4 $ &  $ 64.1 $ &  $ 62.6 $ &  $ 68.6 $ \\
% FairTPE &   $ 55.7 $ &  $ 76.9 $ &  $ 59.2 $ &  $ 80.0 $ \\
%     HB &    $ 68.4 $ &  $ 32.3 $ &  $ 68.4 $ &  $ 35.2 $ \\
%     RS &    $ 67.8 $ &  $ 42.2 $ &  $ 68.1 $ &  $ 45.0 $ \\
%   TPE &    $ 68.7 $ &  $ 30.5 $ &  $ 68.5 $ &  $ 33.7 $ \\

\bottomrule
\\
\end{tabular}